\documentclass[lettersize,journal]{IEEEtran}
\usepackage{amsmath,amsfonts}
\usepackage{algorithmic}
\usepackage{algorithm}
\usepackage{array}
\usepackage[caption=false,font=normalsize,labelfont=sf,textfont=sf]{subfig}
\usepackage{textcomp}
\usepackage{stfloats}
\usepackage{url}
\usepackage{verbatim}
\usepackage{graphicx}
\usepackage{cite}
\usepackage{amssymb}
\usepackage{mathtools}
\usepackage{amsthm}
\usepackage[table]{xcolor}
\usepackage{tabularx}
\usepackage{multicol}
\usepackage{multirow}
\usepackage{booktabs}
\definecolor{grey}{RGB}{230,230,230}
\usepackage{hyperref}

\begin{document}

\title{Watching Synthetic Videos: Aligning Cross-modal Representations with  Visual Synthesis for Zero-shot Video Captioning}

\author{Liangyu Fu\IEEEauthorrefmark{1}, Junbo Wang\IEEEauthorrefmark{1}, Yuke Li\IEEEauthorrefmark{2}, Ya Jing, Xuecheng Wu, Zhiyong Wang
\thanks{\IEEEauthorrefmark{2} \textit{Corresponding author: Yuke Li.}}

\thanks{\IEEEauthorrefmark{1} Both authors contributed equally to this work.}

\thanks{Liangyu Fu, Junbo Wang, and Yuke Li are with the School of Software, Northwestern Polytechnical University, Xi'an 710129, China (e-mail: lyfu@mail.nwpu.edu.cn; jbwang@nwpu.edu.cn; liyuke@nwpu.edu.cn).}


\thanks{Ya Jing is with the School of Information Science and Technology, Beijing University of Technology, Beijing 100124, China (email: jingya004@126.com).}

\thanks{Xuecheng Wu is with the School of Computer Science and Technology, Xi'an Jiaotong University, Xi'an 710049, China (e-mail: wuxc3@stu.xjtu.edu.cn).}

\thanks{Zhiyong Wang is with the School of Computer Science, The University of Sydney, NSW 2006, Australia (e-mail: zhiyong.wang@sydney.edu.au).}

}

\markboth{Journal of \LaTeX\ Class Files,~Vol.~14, No.~8, August~2021}%
{Shell \MakeLowercase{\textit{et al.}}: A Sample Article Using IEEEtran.cls for IEEE Journals}


\maketitle

\begin{abstract}
Text-only training is a popular paradigm in zero-shot video captioning, where the video distribution is not available to the model during training, leading to a cross-modal gap between the training (text-only) and the inference (video-only). Previous works attempt to bridge the gap through simple linear transformations. However, the inherent gap between text and video makes cross-modal representation space alignment insufficient, resulting in inaccurate sentences. To address this issue, we propose a novel zero-shot video captioning framework (WSV) consisting of two training stages, which first generates corresponding synthetic video latent representations via a pretrained text-to-video generation model. To strengthen the fidelity of the latent representations, we propose a polisher capable of bridging the gap between real and synthetic video distributions. Subsequently, we design a prompter that conditions GPT-2 on the polished latent representations to generate the captions in the second training stage. During inference, an input video is encoded by a pretrained 3D Causal VAE and then fed directly into the prompter, which in turn guides GPT-2 to produce the final caption. Experimental results conducted on MSVD, MSR-VTT, and VATEX datasets demonstrate that our proposed method achieves scores of 52 and 95.7 on the B@4 and CIDEr metrics, respectively.
\end{abstract}


\section{Introduction}
\IEEEPARstart{V}{ideo} captioning aims to describe the video in one sentence. Traditional methods rely on strict `video-text pairs' for strongly supervised learning, which not only has extremely high requirements for the scale and quality of the data, but also make it difficult for the model to describe out-of-domain activities that have not appeared during the training process. Therefore, \cite{wang2019learning} proposed the zero-shot video captioning task to overcome the constraints of data and domain. The key to this task is to ensure that the model is effectively trained when out-of-domain data is not available.

Previous works~\cite{12li2023decap,41zhang2024connect,7lee2024ifcap,17liu2024improving,ma2025retta,nukrai-etal-2022-text}, including zero-shot image captioning, have generally adopted the text-only training paradigm, utilizing text data exclusively during training. These methods performed a series of processing or transformations on the text data during training (such as MLP, Gaussian mapping, and Transformers) to make the latent representation obtained from the text conform as closely as possible to the visual distribution in latent space, allowing the text generation component to utilize the visual representations during inference directly.

\begin{figure}[t]
    \centering
    \includegraphics[width=\linewidth]{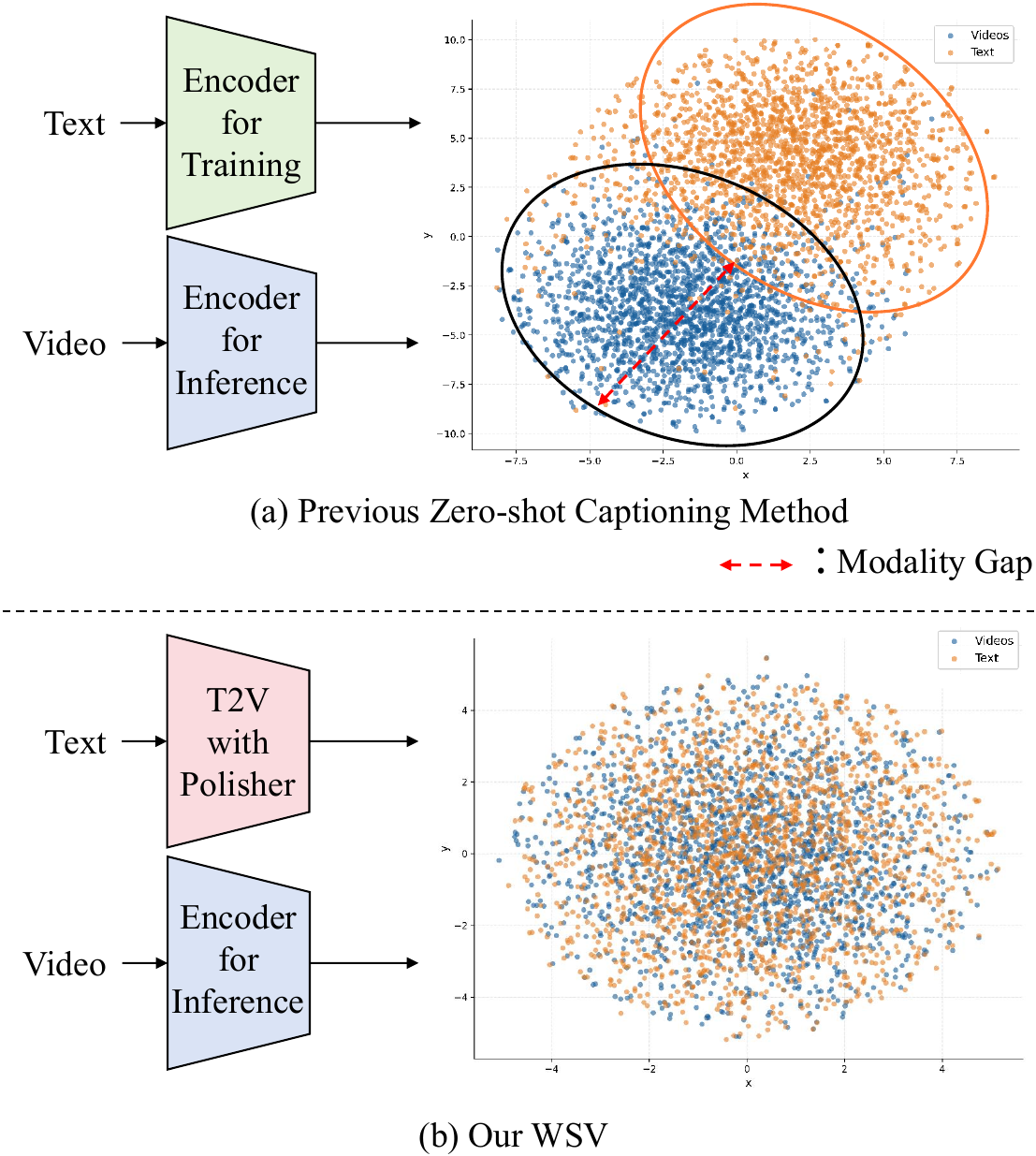}
    \caption{The comparison between (a) Previous Zero-shot Captioning Method and (b) Our WSV. `T2V' denotes the text-to-video model. The t-SNE is calculated based on the representations of real videos and text.}
    \label{fig:intro}
\end{figure}

However, as shown in Fig.~\ref{fig:intro} (a), textual distribution and visual distribution are inherently very different. The latent representation obtained from text can not align well with the visual distribution in latent space via simple linear transformation. This modality gap between vision and language leads to the loss/misalignment of semantic information (especially temporal semantic information) between training (text-only) and inference (video), preventing the subsequent text generation component from receiving complete semantic information and ultimately affecting the quality of the textual descriptions.

To address this issue, as shown in Fig.~\ref{fig:intro} (b), we propose a zero-shot video captioning method that enables model to \textbf{W}atch \textbf{S}ynthetic \textbf{V}ideos (WSV) under the text-only training paradigm, thereby bridging the modality gap between training and inference. Specifically, during training, we follow the text-only training paradigm, and leverage the powerful capabilities of current text-to-video models to synthesis video latent representations from text corpus. To ensure the fidelity of the synthetic video latent representations, we design a distribution aligner that bridges the distribution of real and synthetic videos. Next, we design a prompter to make prompts based on the latent representations. Finally, to maintain consistency with previous work~\cite{ma2025retta} settings and ensure the fairness of the evaluation, we use GPT-2 to generate the captions. During inference, we directly encode the video and input it into the prompter. GPT-2 then generates captions based on the prompts. The WSV align the modality and avoids the loss of visual semantics during training, thereby improving caption quality.

In summary, our contributions are as follows:

\begin{itemize}
    \item We propose a novel zero-shot video captioning framework that leverages state-of-the-art text-to-video generation models to synthesize visual content from text corpus, thereby bridging the modal gap between visual and textual latent representations and reducing the loss of semantic information.
    \item We propose a distribution aligner that conforms the generated video content to the distribution of real-world videos, thus improving the quality of visual information. 
    \item We propose a prompter that makes high-quality prompts for the GPT-2 based on the visual representation.
    \item Extensive experiments on MSVD, MSR-VTT and VATEX demonstrate that our method achieves state-of-the-art zero-shot video captioning performance, reaching 52 and 95.7 on the B@4 and CIDEr metrics, respectively.
\end{itemize}

\section{Related Work}
\label{sec:Related Work}

\subsection{Video Captioning}

Video captioning aims to generate the textual description of a video in one sentence. Compared to image captioning, it needs to handle additional temporal and spatially complex information. Early work~\cite{31venugopalan2015sequence,1bin2016bidirectional,11li2022graph,37yao2015describing, icocap, man} focused on techniques for visual encoding and text decoding within an encoder-decoder framework. Since the emergence of Transformer~\cite{29vaswani2017attention}, numerous research~\cite{38ye2022hierarchical,14lin2022swinbert} efforts have utilized it to study more advanced visual modeling and text generation, greatly improving model performance. Recently, works such as Video-LLaMA~\cite{40zhang2023video} and MA-LMM~\cite{5he2024ma} have begun to utilize large language models (LLMs) as subtitle generators to describe video, fully leveraging the powerful generalization capabilities.

\subsection{Zero-shot Image and Video Captioning}

Zero-shot image captioning aims to address the domain generalization limitation of models. Previous methods typically generated textual descriptions autoregressively according to an encoder-decoder architecture~\cite{20nukrai2022text,26su2022language,27tewel2022zerocap}. The encoder part was basically implemented by CLIP~\cite{22radford2021learning}, and the decoder part was usually implemented by a Transformer-based language model. They calculate visual and text similarity once after each time step, which disrupts the coherence of the text and the overall semantic alignment between image and text. Furthermore, they still require data in the form of image-text pairs during training, which is quite difficult for zero-shot tasks. To address the issues, \cite{12li2023decap} and~\cite{32wang2023association} proposed the text-only training paradigm that uses only text as training data during the training process. This overcomes the lack of data and promotes the overall semantic alignment of sentences with images. Subsequent methods~\cite{41zhang2024connect,7lee2024ifcap,17liu2024improving} followed the text-only training paradigm, but the modality gap between vision and language remained unresolved. Meanwhile, these zero-shot methods are designed for image captioning. The approach of using pooled video frames for zero-shot video captioning has been proven to be ineffective. This often results in the loss of important information in the video, including temporal information, which seriously damages the quality of the generated captions.

Similar to the challenges faced by image captioning, the video captioning methods described above rely on training with large-scale video-text pairs and lack the ability to generalize to out domains. However, there is little research in zero-shot video captioning. \cite{wang2019learning} propose a principled Topic-Aware Mixture of Experts (TAMoE) model for zero-shot video captioning, which learns to compose different experts based on different topic embeddings, implicitly transferring the knowledge learned from seen activities to unseen ones. \cite{ma2025retta} propose a novel zero-shot video captioning framework named Retrieval-Enhanced Test-Time Adaptation (RETTA), which takes advantage of existing pre-trained large-scale vision and language models to directly generate captions with test-time adaptation. Due to the relatively strict limitations on the training data, the performance of these methods is generally poor. However, if the text-only training paradigm is applied to zero-shot video captioning to address the data demand, the modality gap between training and inference is a key issue that must be addressed.

\begin{figure*}[t]
    \centering
    \includegraphics[width=\linewidth]{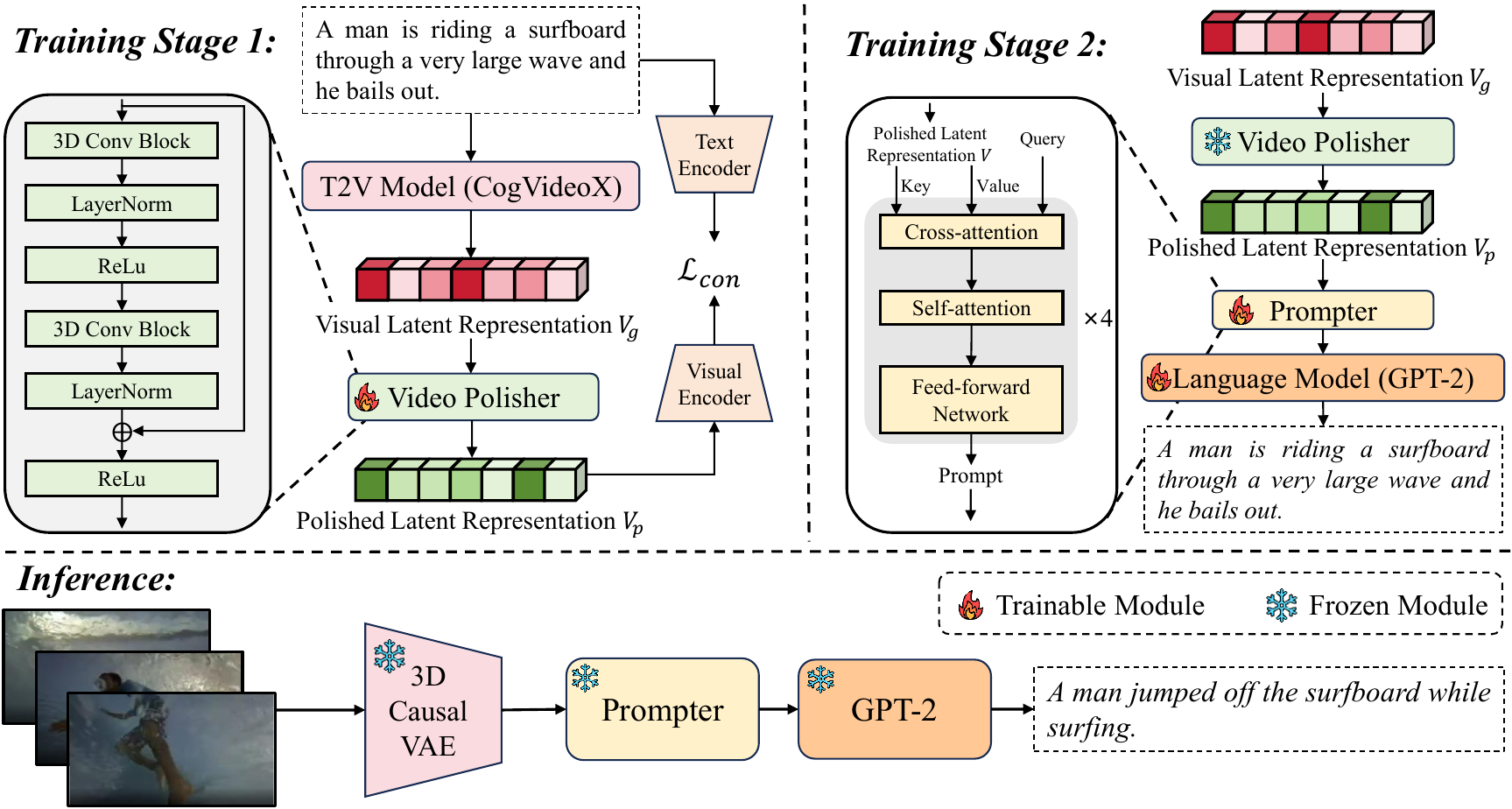}
    \caption{The overall framework of WSV. Regular text denotes annotated textual descriptions from the corpus, while italic text denotes generated textual descriptions. The Text Encoder and Video Encoder in Training Stage 1 are pre-trained CLIP4Clip~\cite{luo2022clip4clip}. The 3D Causal VAE is the pre-trained video encoder from CogVideoX~\cite{yang2024cogvideox}.}
    \label{fig:or}
\end{figure*}

\section{Methodology}
\label{sec:Mthods}

To align the modality between training (text-only) and inference (video), we propose a novel zero-shot video captioning framework that trains a video captioner by \textbf{W}atching \textbf{S}ynthetic \textbf{V}ideos (WSV). The overall framework is illustrated in Fig.~\ref{fig:or}. Specifically, WSV is comprised of three distinct stages: (1) Visual Latent Synthesis and Polishing, (2) Captioner Training, and (3) Inference. In stage 1, we leverage a pre-trained Text-to-Video (T2V) model as a powerful generative prior, synthesizing visual latent representations directly from the text-only corpus. Furthermore, we designed a polisher implemented by a 3D convolutional neural network to improve the quality of the latent representations and reduce potential errors. In stage 2, our captioner is trained on these synthesis visual latent representations, ensuring the model learns from an inherently visual data distribution, thereby eliminating the modality gap in inference. For the inference, we replace the T2V model and polisher with the pre-trained video encoder from the T2V model, it encodes the video and inputs the latent representations to the prompter. Finally, GPT-2 generates the caption based on the prompts.


\subsection{Stage 1: Visual Latent Synthesis and Polishing}

The key challenge of the text-only training paradigm lies in aligning the text distribution in the latent space with the visual distribution. Previous projection-based methods were insufficient to ensure that the text representation during training conformed to the visual distribution. Therefore, the visual distribution was invisible during training, leading to model unfamiliarity with it during inference—a phenomenon known as modality gap. With the development of text-to-video models, converting the text distribution to the visual distribution has become feasible. We leverage this advantage to enable the zero-shot model to learn the true visual distribution during training, thereby eliminating the modality gap.

Specifically, given a large text-only corpus $\mathcal{T} = \{t_i\}_{i=1}^N$ (e.g., from MSVD, MSR-VTT, and VATEX), we feed each caption $t_i$ into CogVideoX~\cite{yang2024cogvideox}, $\mathcal{G}_{T2V}$. Crucially, instead of decoding the full, pixel-space video, which is computationally prohibitive, we intercept the generation process to extract the final latent representation $V_g \in \mathbb{R}^{C \times F \times H \times W}$. This process can be expressed as:

\begin{equation}
    V_{g}^i = \mathcal{G}_{T2V}(t_i).
\end{equation}

This latent $V_{g}^i$ is the output of the CogVideoX before it enters the final VAE decoder. It is a compact representation of the synthesized video, capturing rich spatio-temporal structure.

This process results in a new pseudo-dataset $\mathcal{D}_L = \{(V_{g}^i, t_i)\}_{i=1}^N$, where $V_{g}^i$ serves as the generative visual surrogate for the caption $t_i$.

As shown in Fig.~\ref{fig:tsne_mismatch}, Due to the latent $V_g$ from Stage 1 is synthesized by the CogVideoX, this distribution may not perfectly match the distribution of latents produced by the encoder of the same VAE when processing real videos. To mitigate this potential mismatch and refine the visual quality of our surrogates, we introduce a polisher $\mathcal{A}_{\phi}$, parameterized by $\phi$. It is a lightweight 3D Residual Convolutional Network that transforms $V_g$ into an high-fidelity latent $V_p$, while strictly preserving its 4D spatio-temporal dimensions, the structure of polisher is shown in Fig.~\ref{fig:or}. This process can be expressed as: 

\begin{equation}
    V = \mathcal{A}_{\phi}(V_g).
\end{equation}

To train the polisher, we introduce the contrastive objective. The role of $\mathcal{A}_{\phi}$ is to produce a latent $V_p$ that, when decoded, is semantically aligned with the original source text $t_i$. We feed the latent $V_p$ through the CogVideoX's VAE decoder ($\mathcal{D}_{VAE}$) to get a pixel-space video $\hat{v}$. This video $\hat{v}$ is then passed through a pre-trained, frozen video-text model (e.g., CLIP4Clip~\cite{luo2022clip4clip}'s video encoder, $\mathcal{E}_{CLIP\_V}$) to obtain a visual embedding $f_v$. The corresponding caption $t_i$ is encoded via the text encoder $\mathcal{E}_{CLIP\_T}$ to get $f_t$. The process can be expressed as: 

\begin{equation}
    f_v = \mathcal{E}_{CLIP\_V}(\mathcal{D}_{VAE}(V_p)),
\end{equation}

\begin{equation}
    \quad f_t = \mathcal{E}_{CLIP\_T}(t_i).
\end{equation}

The polisher's parameters $\phi$ (and those of the temporal modeler within $\mathcal{E}_{CLIP\_V}$) are optimized using a symmetric cross-entropy loss $\mathcal{L}_{con}$ over a batch of $B$ pairs. Let $s(f_v, f_t) = \text{sim}(f_v, f_t) / \tau$ denote the scaled cosine similarity with a temperature $\tau$:

\begin{equation}
\begin{split}
    \mathcal{L}_{con} = - \frac{1}{2B} \sum_{i=1}^{B} \left( \log \frac{\exp(s(f_{v,i}, f_{t,i}))}{\sum_{j=1}^{B} \exp(s(f_{v,i}, f_{t,j}))} \right. \\
    \left. +\log \frac{\exp(s(f_{t,i}, f_{v,i}))}{\sum_{j=1}^{B} \exp(s(f_{t,i}, f_{v,j}))} \right)
\end{split}
\end{equation}

The video-text encoder (CLIP) is used only as a supervisory signal in this stage to `teach' the polisher $\mathcal{A}_{\phi}$ how to produce semantically coherent latent videos. As shown in Fig.~\ref{fig:tsne_aligned}, compared to direct synthesis, the representations processed by the polisher have a better alignment with the real video representations.

\subsection{Stage 2: Captioner Training}


With the frozen, pre-trained polisher $\mathcal{A}_{\phi}$ in hand, we now train our primary video captioning model. This model consists of two components: a novel prompter $\mathcal{P}_{\theta}$, parameterized by $\theta$, and a pre-trained language model $\mathcal{L}_{\psi}$ (e.g., GPT-2~\cite{radford2019gpt2}), parameterized by $\psi$. This stage operates without any CLIP encoders or VAE decoders. The entire pipeline is trained in the latent space.

We use our pseudo-dataset $\mathcal{D}_L$ again. For each pair $(V_{g}, t_i)$, the generative latent $V_{g}$ is passed through the frozen polisher. Then the resulting high-fidelity latent $V_p$ is fed into the prompter $\mathcal{P}_{\theta}$ to produce prompt embeddings $P_v$, the overall process can be expressed as:

\begin{equation}
    P_v = \mathcal{P}_{\theta}(V_p)
\end{equation}

As illustrated in Fig.~\ref{fig:or}, the prompter $\mathcal{P}_{\theta}$ is designed to map the 4D spatio-temporal latent $V_p$ into a sequence of soft prompt embeddings. It first uses a 3D CNN to encode $V_p$ into a 1D sequence of features $F_v \in \mathbb{R}^{L \times D_{LM}}$.

\begin{equation}
    F_v = \text{Flatten}(\text{CNN}_{3D}(V_p))
\end{equation}

These features (as Key and Value) are then fused with a set of learnable queries $Q_p$ (as Query) using a block which combines cross-attention and self-attention.

\begin{equation}
    P_v = \text{SelfAttn}(\text{CrossAttn}(Q_p, F_v))
\end{equation}

The output is a fixed-length sequence of prompt embeddings $P_v \in \mathbb{R}^{N_p \times D_{LM}}$, where $N_p$ is the number of prompt tokens and $D_{LM}$ is the LM's embedding dimension. These prompt embeddings $P_v$ are prepended to the word embeddings $E_{t_i}$ (from the LM's embedding layer $\mathcal{E}_{LM}$) of the ground-truth caption $t_i$.

\begin{table*}[ht]
    \centering
    \caption{Supervised and Zero-shot Video captioning model performance on MSR-VTT, MSVD, and VATEX datasets. - denotes that the data is not given in the corresponding literature. The bold number denotes the best results among all zero-shot methods.}
    \begin{tabular}{l|cccc|cccc|cccc}
    \toprule
    \multirow{2}{*}{Method} & \multicolumn{4}{c|}{MSR-VTT} & \multicolumn{4}{c|}{MSVD} & \multicolumn{4}{c}{VATEX}\\
    \cmidrule(lr){2-5} \cmidrule(lr){6-9} \cmidrule(lr){10-13}
    & B@4 & M & R & C & B@4 & M & R & C & B@4 & M & R & C\\
    \midrule
    \rowcolor{grey}\multicolumn{13}{l}{\textit{{Supervised}}}\\
    VNS-GRU~\cite{chen2020delving} & 45.3 & 29.9 & 63.4 & 53.0 & 66.5 & 42.1 & 79.7 & 121.5 & - & - & - & -\\
    SemSynAN~\cite{perez2021improving} & 46.4 & 30.4 & 64.7 & 51.9 & 64.4 & 41.9 & 79.5 & 111.5 & - & - & - & -\\
    HMN~\cite{38ye2022hierarchical} & 43.5 & 29.0 & 62.7 & 51.5 & 59.2 & 37.7 & 75.1 & 104.0 & 33.2 & 22.7 & 49.3 & 51.6\\
    \midrule
    \rowcolor{grey}\multicolumn{13}{l}{\textit{{Zero-shot}}}\\
    ZeroCap~\cite{27tewel2022zerocap} & 2.3 & 12.9 & 30.4 & 5.8 & 2.9 & 16.3 & 35.4 & 9.6 & - & - & - & -\\
    MAGIC~\cite{26su2022language} & 5.5 & 13.3 & 35.4 & 7.4 & 6.6 & 16.1 & 40.1 & 14.0 & - & - & - & -\\
    EPT~\cite{tewel2022zero} & 3.0 & 14.6 & 27.7 & 11.3 & 3.0 & 17.8 & 31.4 & 17.4 & - & - & - & -\\
    Video-LLaMA~\cite{40zhang2023video} & 4.9 & 16.8 & 25.3 & 2.3 & 4.0 & 15.1 & 21.6 & 1.3 & 4.3 & 16.3 & 21.8 & 3.8\\
    ZS-CapCLIP~\cite{22radford2021learning} & 4.0 & 15.0 & 31.0 & 5.0 & - & - & - & - & - & - & - & -\\
    MultiCapCLIP~\cite{yang-etal-2023-multicapclip} & 13.3 & 19.5 & 43.3 & 15.5 & - & - & - & - & - & - & - & -\\
    DeCap-COCO~\cite{12li2023decap} & 14.7 & 20.4 & - & 18.6 & - & - & - & - & 13.1 & 15.3 & - & 18.7\\
    LLaVa~\cite{liu2023visual} & - & - & - & 16.9 & - & - & - & - & 4.3 & 16.3 & 21.8 & 3.8\\
    AuroraCap-7B~\cite{chai2024auroracap} & 21.0 & 23.9 & 49.5 & 33.1 & - & - & - & - & 18.4 & 19.0 & 40.8 & 33.8\\
    RETTA~\cite{ma2025retta} & 14.0 & 19.3 & 42.2 & 24.3 & 23.3 & 28.5 & 56.4 & 49.8 & 11.4 & 16.3 & 32.6 & 23.8\\
    \midrule
    \textbf{WSV (Ours)} & \textbf{33.9} & \textbf{27.7} & \textbf{55.7} & \textbf{45.5} & \textbf{52} & \textbf{38.7} & \textbf{72.3} & \textbf{95.7} & \textbf{26.1} & \textbf{18.6} & \textbf{40.2} & \textbf{35.5}\\
    \bottomrule
    \end{tabular}
    \label{tab:ce}
\end{table*}

\begin{figure*}[t]
    \centering
    \includegraphics[width=\linewidth]{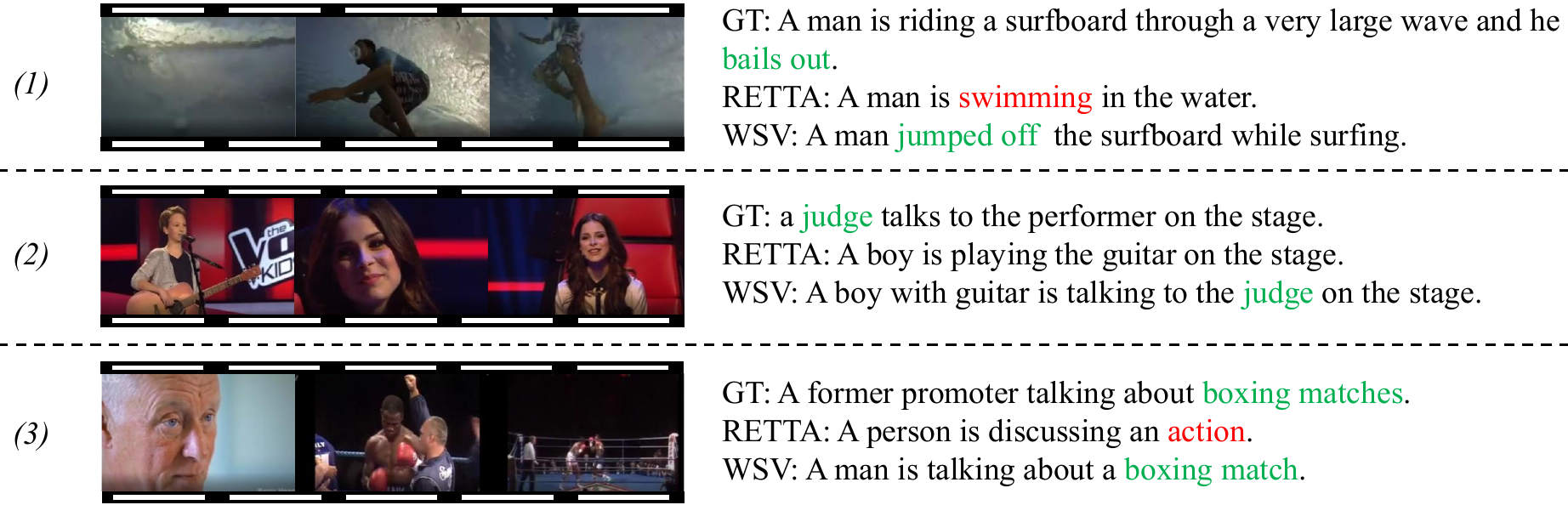}
    \caption{Three case studies from the MSR-VTT dataset. Each sample contains textual descriptions from two video captioning models and one labeled description. Red indicates incorrect content, and green indicates content that matches the labeled caption.}
    \label{fig:case}
\end{figure*}

\begin{equation}
    E_{in} = [P_v ; E_{t_i}] = [P_v ; \mathcal{E}_{LM}(t_i)]
\end{equation}

The full sequence $E_{in}$ is fed into the LM $\mathcal{L}_{\psi}$. The model is trained to predict the next token in the caption $t_i$ using a standard auto-regressive cross-entropy loss:

\begin{equation}
    \mathcal{L}_{CE} = - \sum_{j=1}^{|t_i|} \log P(t_{i,j} | E_{in, <j}, \psi)
\end{equation}

During this stage, we update the parameters of both the prompter, $\theta$, and the GPT-2, $\psi$, to minimize this loss:

\begin{equation}
    \theta^*, \psi^* = \arg \min_{\theta, \psi} \mathcal{L}_{CE}
\end{equation}

\subsection{Inference}

Given a real video $v_{real}$, we first pass it through the frozen VAE Encoder $\mathcal{E}_{VAE}$ (from the CogVideoX) to obtain its real latent representation. The real latent $V_{g-real}$ is then processed by the prompter The entire visual-to-prompt encoding can be expressed as:

\begin{equation}
    P_{v-real} = \mathcal{P}_{\theta}(\mathcal{E}_{VAE}(v_{real}))
\end{equation}

LM Generation: The LM $\mathcal{L}_{\psi}$ receives only the prompt embeddings $P_{v-real}$ as input and auto-regressively generates the final caption token by token:

\begin{equation}
    t_{caption} = \mathcal{L}_{\psi}^{\text{generate}}(P_{v-real})
\end{equation}

By training our captioner on generated visual surrogates that have passed through the prompter, we ensure it is perfectly prepared to handle real visual latents from the VAE encoder at test time. This unified pipeline, trained on a synthetic visual domain, successfully bridges the modality gap.

\section{Experiments}
\label{sec:exprs}

\subsection{Datasets}
Following state-of-the-art zero-shot video captioning work, we conducted experiments on the MSR-VTT, MSVD, and VATEX datasets. 

MSR-VTT~\cite{xu2016msr} contains 10,000 video clips covering 20 categories. Each clip has 20 English annotations. The standard split includes 6513 training videos, 497 validation videos, and 2990 test videos. MSVD~\cite{chen2011msvd} contains 1970 video clips. Each clip has roughly 40 ground-truth captions. The standard split includes 1200 training videos, 100 validation videos, and 670 test videos. VATEX~\cite{wang2019vatex} contains 34,991 videos with 10 English annotations. The standard split includes 25,910 training videos, 3000 validation videos, and 6000 test videos. For WSV, text from these three datasets was mixed to form the training text corpus. The experimental results were derived from the model's cross-domain inference on the test sets of each dataset.

\subsection{Metrics}

We employ multiple standard metrics in video captioning tasks, including BLEU (B)~\cite{papineni2002bleu}, METEOR (M)~\cite{denkowski2014meteor}, ROUGE-L (R)~\cite{lin2004rouge} and CIDEr (C)~\cite{vedantam2015cider}. BLEU@4 measures n-gram precision and reflects whether the generated caption contains locally accurate phrase patterns. METEOR evaluates word-level matching with stronger consideration of recall, making it useful for measuring semantic adequacy. ROUGE-L focuses on the longest common subsequence between generated and reference captions, which reflects sentence-level structural consistency. CIDEr measures consensus with human annotations and is particularly important for captioning because it rewards visually specific and informative descriptions. We report all four metrics to provide a comprehensive evaluation of generation quality.

\subsection{Implementation Details}

For CogVideoX, we use the CogVideoX-5B model on Hugging Face. Following prior works~\cite{7lee2024ifcap,12li2023decap, ma2025retta}, we choose the standard version of GPT-2~\cite{radford2019gpt2}. We optimize all trainable parameters with the AdamW~\cite{loshchilov2017decoupled} optimizer. We use distinct learning rates for the two stages: 1e-4 for Stage 1 and a smaller 2e-5 for Stage 2 (Prompter + GPT-2). We implement our framework with PyTorch 2.0.0. All experiments are conducted on 8 NVIDIA Tesla V100 GPUs.

\subsection{Performance Comparison}

\textbf{Quantitative Analysis}: 
As shown in Table~\ref{tab:ce}, we compare WSV with representative supervised and zero-shot video captioning methods on MSR-VTT, MSVD, and VATEX. Although supervised models are trained with paired video-text data, WSV achieves highly competitive performance under the more challenging zero-shot setting. On MSR-VTT, WSV obtains 33.9 on B@4, 27.7 on METEOR, 55.7 on ROUGE-L, and 45.5 on CIDEr. These results are substantially higher than previous zero-shot methods. Compared with the best previously reported zero-shot results on MSR-VTT, WSV improves B@4 from 21.0 to 33.9, METEOR from 23.9 to 27.7, ROUGE-L from 49.5 to 55.7, and CIDEr from 33.1 to 45.5. The 12.4-point gain on CIDEr indicates that WSV generates captions that better match human reference descriptions and contain more discriminative visual content.

On MSVD, WSV reaches 52.0 on B@4, 38.7 on METEOR, 72.3 on ROUGE-L, and 95.7 on CIDEr. Compared with RETTA, which reports results on all three datasets, WSV improves B@4 by 28.7, METEOR by 10.2, ROUGE-L by 15.9, and CIDEr by 45.9. This large margin shows that the proposed synthetic visual training pipeline is especially effective on datasets with dense annotations, where the model must produce semantically precise and human-consistent descriptions. On VATEX, WSV obtains 26.1 on B@4, 18.6 on METEOR, 40.2 on ROUGE-L, and 35.5 on CIDEr. It achieves the highest B@4 and CIDEr among the listed zero-shot methods, improving the previous best scores by 7.7 and 1.7, respectively. These results demonstrate that WSV consistently narrows the modality gap between text-only training and real-video inference. Instead of relying only on text-space projection or test-time prompt optimization, WSV exposes the captioner to visual latent distributions during training, leading to stronger cross-dataset captioning performance.

\textbf{Qualitative Analysis}: 
We present three test samples of WSV and previous zero-shot methods on MSR-VTT, and the results are shown in Fig.~\ref{fig:case}. Specifically, the video semantics of Sample 1 mainly consist of `man, surfboard, and bails out. RETTA's caption incorrectly interprets `surfing' as `swimming' and omits the element of `bails out'. On the other hand,  WSV successfully understood all three key semantics. In Sample 2, `talking to the judge' is a key semantic element in the video. RETTA only understood the boy and ignored the content related to the `judges'. In Sample 3, RETTA broadly interprets `boxing matches' as `action movies', while WSV accurately points out the key content of `boxing matches' by combining the context. In general, sample 1 and 3 demonstrate WSV's more accurate generation capability, while sample 2 demonstrates WSV's more comprehensive understanding capability.

\begin{table*}[t]
    \centering
    \caption{Ablation study for video captioning on MSR-VTT, MSVD, and VATEX datasets. The bold number denotes the best results among all methods.}
    
    \begin{tabular}{lcccc|cccc|cccc}
    \toprule
        \multirow{2}{*}{Method} & \multicolumn{4}{c|}{MSR-VTT} & \multicolumn{4}{c|}{MSVD} & \multicolumn{4}{c}{VATEX}\\
        \cmidrule(lr){2-5} \cmidrule(lr){6-9} \cmidrule(lr){10-13}
        & B@4 & M & R & C & B@4 & M & R & C & B@4 & M & R & C \\
        \midrule
        WSV & \textbf{33.9} & \textbf{27.7} & \textbf{55.7} & \textbf{45.5} & \textbf{52.0} & \textbf{38.7} & \textbf{72.3} & \textbf{95.7} & \textbf{26.1} & \textbf{18.6} & \textbf{40.2} & \textbf{35.5} \\
        \textit{w/o} Synthetic Visual & 31 & 25.4 & 49.2 & 39 & 49.5 & 34.3 & 62.1 & 82.4 & 22.8 & 15.4 & 31.5 & 27\\
        \textit{w/o} Polisher & 33.1 & 25.7 & 50.4 & 40.4 & 50.7 & 32.4 & 68.9 & 91.6 & 25 & 16.1 & 35.7 & 29.9\\
        \textit{w/o} Prompter & 31.5 & 22.3 & 49.2 & 39.7 & 49.9 & 36.7 & 68.4 & 82.7 & 23.1 & 15.7 & 37.8 & 29.1\\
    \bottomrule
    \end{tabular}
    
    \label{tab:ae}
\end{table*}

    
    

    
    

\subsection{Ablation Study}

    
    

To further analyze the contribution of each component, we conduct ablation experiments on the visual synthesis module, the polisher, the prompter, the number of CNN blocks in the polisher, the number of attention blocks in the prompter, and the CLIP encoder used for Stage 1 supervision. Unless otherwise stated, each ablation changes only the target component and keeps the remaining WSV configuration unchanged.

\begin{figure}[t]
    \centering
    \includegraphics[width=\linewidth]{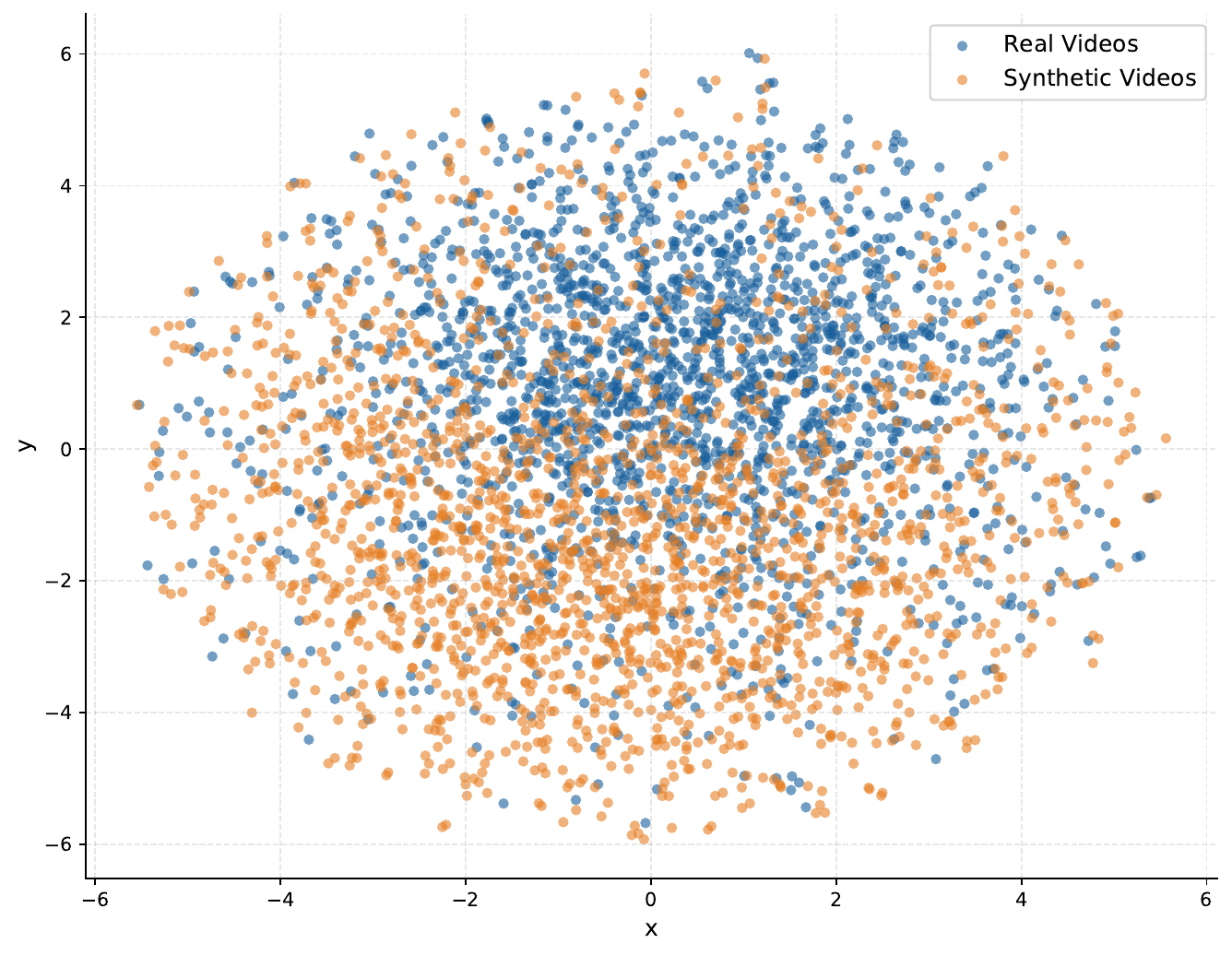}
    \caption{The t-SNE of the real video representation and the synthetic video representation from MSR-VTT~\cite{xu2016msr}. Although these two distributions have achieved a certain degree of alignment, the visible distribution gap (e.g., real videos leaning upper-right vs. synthetic videos leaning lower-left) highlights the mismatch, which motivates the design of our Polisher module.}
    \label{fig:tsne_mismatch}
\end{figure}

\begin{figure}[t]
    \centering
    \includegraphics[width=\linewidth]{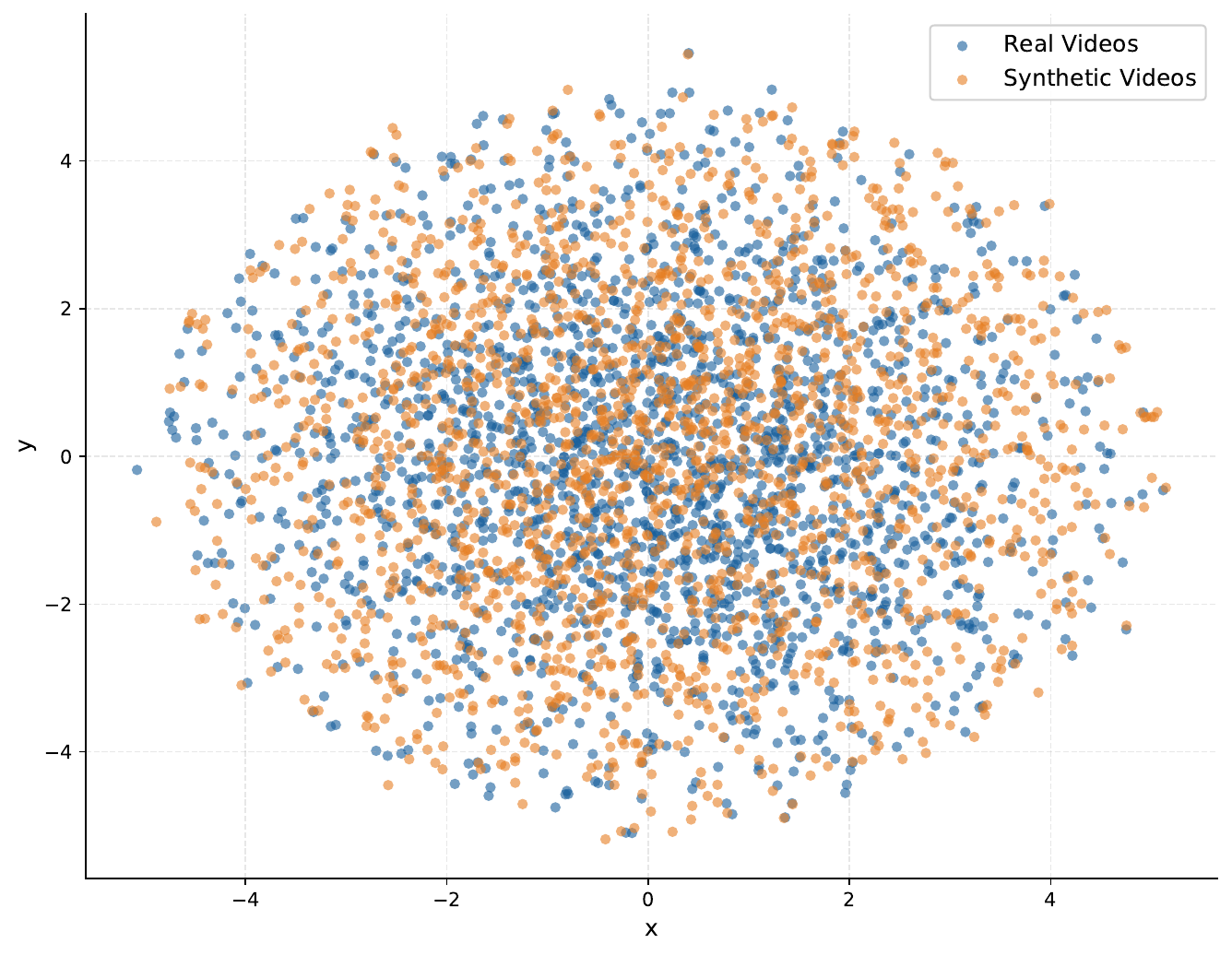}
    \caption{The t-SNE of the real video representation and the synthetic video representation optimized by polisher from MSR-VTT~\cite{xu2016msr}.}
    \label{fig:tsne_aligned}
\end{figure}

\begin{figure}[t]
    \centering
    \includegraphics[width=\linewidth]{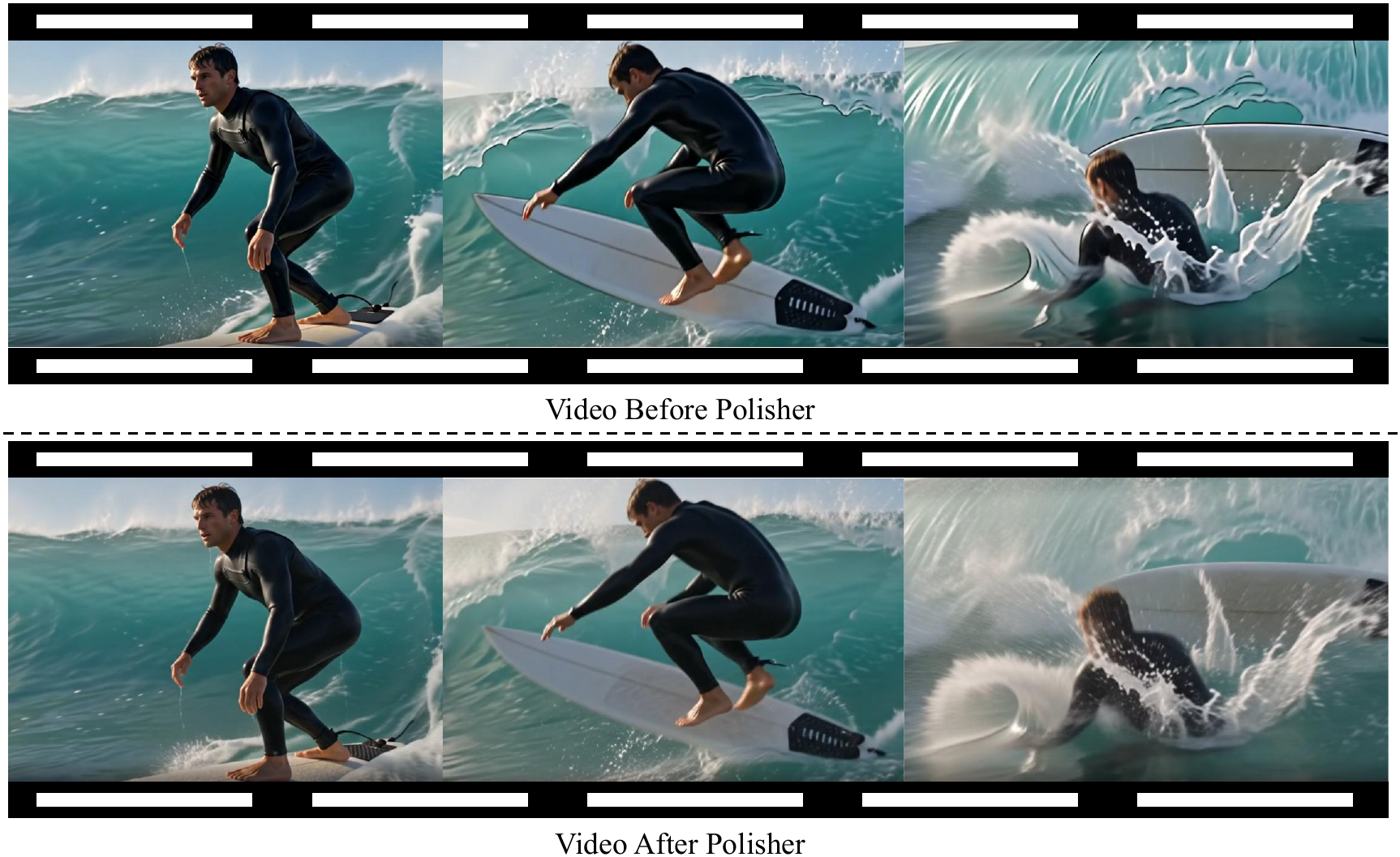}
    \caption{Comparison of a video sample before and after the Polisher.}
    \label{fig:pcase}
\end{figure}

\textbf{The Impact of Polisher.} Table~\ref{tab:ae} shows the ablation experiment results for the polisher. ‘\textit{w/o} Polisher’ indicates that the polisher was not used in the Stage 1, and the synthesized video latent representation was directly input into the Prompter for the second stage training. The experimental results show that the polisher improves the quality of the video latent representation generated by the T2V model, and this quality improvement also affects the final text generation. Fig.~\ref{fig:pcase} shows the changes in the video before and after the Polisher processing. Before the processing, the objects in the video lacked a sense of reality. After the processing, the contours and details of the objects became clearer and more realistic. Furthermore, Fig.~\ref{fig:tsne_mismatch} and Fig.~\ref{fig:tsne_aligned} illustrate the changes in the distribution alignment between the real and synthetic video representations. We can observe that, compared to before Polisher processing (Fig.~\ref{fig:tsne_mismatch}), the distribution of the synthetic video representation after Polisher processing (Fig.~\ref{fig:tsne_aligned}) has a higher alignment with the real video representation.

\textbf{The Impact of Synthetic Visual.} Table~\ref{tab:ae} presents the ablation experiment results for synthetic visual content. ‘\textit{w/o} Synthetic Visual’ indicates that the T2V model in WSV was replaced with the CLIP text encoder, without using Polisher. The Prompter was used during the Stage 2 training, and the CLIP video encoder was used during inference. Experimental results show that synthetic visual content significantly enhances the performance of zero-shot video captioning through modality alignment in the latent space, demonstrating improved quality across multiple metrics on all three datasets.

\textbf{The Impact of Prompter.} Table~\ref{tab:ae} shows the ablation experiment results for the Prompter. ‘\textit{w/o} Prompter’ indicates that the same configuration as WSV was maintained in the first training phase, but the Prompter was not used in the second training phase; instead, the latent representation produced in the first phase was directly input into GPT-2 as a cue. The experimental results show that the Prompter is necessary for improving the quality of text generation, effectively helping GPT-2 to better understand the latent representation.

    
    

\begin{table}[t]
    \centering
    \caption{Ablation study of the CNN block's number for video captioning on MSR-VTT Datasets. The bold number denotes the best results among all methods.}
    
    \begin{tabular}{ccccc}
    \toprule
        \multirow{2}{*}{Num. of CNN Blocks} & \multicolumn{4}{c}{MSR-VTT}\\
        \cmidrule(lr){2-5}
        & B@4 & M & R & C\\
        \midrule
        1 & 33 & 22.6 & 51.1 & 42.2\\
        2 & \textbf{33.9} & \textbf{27.7} & \textbf{55.7} & \textbf{45.5} \\
        3 & 33.4 & 25.8 & 55.1 & 39\\
        4 & 31 & 21.5 & 52.3 & 40.7\\
    \bottomrule
    \end{tabular}
    
    \label{tab:aepo1}
\end{table}

\textbf{The Impact of the Number of CNN Blocks in Polisher.} Table~\ref{tab:aepo1} shows the impact of CNN blocks' number in the polisher on the final generated quality. Each CNN block contains a 3D CNN block, a layernorm, and a ReLu. Experimental results show that WSV achieves significantly better results when the number of CNN blocks is 2. As the number of CNN blocks increases, the final generated quality of the model tends to decrease. Therefore, we use a configuration of 2 CNN blocks in the polisher.

    
    

\begin{table}[t]
    \centering
    \caption{Ablation study of the Attention block's number for video captioning on MSR-VTT Datasets. The bold number denotes the best results among all methods.}
    
\begin{tabular}{ccccc}
    \toprule
        \multirow{2}{*}{Num. of Attention Blocks} & \multicolumn{4}{c}{MSR-VTT}\\
        \cmidrule(lr){2-5}
        & B@4 & M & R & C\\
        \midrule
        1 & 28.2 & 18.7 & 40.5 & 30.0\\
        4 & \textbf{33.9} & 27.7 & \textbf{55.7} & \textbf{45.5}\\
        8 & 33.9 & \textbf{28.2} & 55.5 & 40.3\\
    \bottomrule
    \end{tabular}
    
    \label{tab:aepro1}
\end{table}

\textbf{The Impact of the Number of Attention Block.} Table~\ref{tab:aepro1} shows the impact of the number of attention blocks in the prompter on the final generation performance of the model. Each attention block contains a cross-attention layer, a self-attention layer, and a feed-forward network. Experimental results show that WSV achieves significantly better results when the number of blocks is 4 or 8. Considering computational costs, we chose to deploy 4 attention blocks in the prompter.

    
    

    
    

\begin{table}[t]
    \centering
    \caption{Ablation study on CLIP Encoder for zero-shot video captioning on MSR-VTT Datasets. The bold number denotes the best results among all methods.}
\begin{tabular}{ccccc}
    \toprule
        \multirow{2}{*}{Method} & \multicolumn{4}{c}{MSR-VTT}\\
        \cmidrule(lr){2-5}
        & B@4 & M & R & C\\
        \midrule
        CLIP4clip & \textbf{33.9} & 27.7 & \textbf{55.7} & 45.5\\
        XCLIP & 33.2 & \textbf{28} & 54.9 & \textbf{47.4}\\
    \bottomrule
    \end{tabular}
    
    \label{tab:clip1}
\end{table}

\textbf{The Impact of the CLIP Encoder.} To investigate the impact of the CLIP encoder on model performance during training stage I, we conducted ablation experiments as shown in Table~\ref{tab:clip1}. We used two CLIP encoders capable of encoding video representations: XCLIP~\cite{ni2022expanding} and CLIP4Clip~\cite{luo2022clip4clip}. The two groups maintained the same configuration except for the CLIP encoder used in the training stage I. Experimental results show that the performance achieved by the two different encoders is essentially the same, demonstrating that the choice of CLIP encoder does not significantly affect model performance.

\begin{table}[t]
    \centering
    \caption{Ablation Study of Language Model's Status on MSR-VTT Dataset}
    
    \begin{tabular}{ccccc}
    \toprule
        \multirow{2}{*}{Method} & \multicolumn{4}{c}{MSR-VTT} \\
        \cmidrule(lr){2-5}
        & B@4 & M & R & C\\
        \midrule
        Frozen & 31.5 & 24.9 & 53.4 & 41.8\\
        Fine-tune & \textbf{33.9} & \textbf{27.7} & \textbf{55.7} & \textbf{45.5}\\
    \bottomrule
    \end{tabular}
    
    \label{tab:status}
\end{table}

\textbf{The Impact of Language Model's Status.} Table~\ref{tab:status} investigates whether the language model should remain frozen or be fine-tuned during the captioner training stage. When GPT-2 is frozen, the prompter must project all visual information into a fixed language embedding space. This setting preserves the original language prior of GPT-2, but it also restricts the model's ability to adapt to the distribution of prompts produced from synthetic visual latents. As a result, the frozen setting obtains 31.5 on B@4, 24.9 on METEOR, 53.4 on ROUGE-L, and 41.8 on CIDEr.

Fine-tuning the language model consistently improves all four metrics. Compared with the frozen setting, the fine-tuned model improves B@4 from 31.5 to 33.9, METEOR from 24.9 to 27.7, ROUGE-L from 53.4 to 55.7, and CIDEr from 41.8 to 45.5. The gains are especially clear on METEOR and CIDEr, with absolute improvements of 2.8 and 3.7, respectively. This indicates that the language model needs to adapt not only to the textual style of the target captions, but also to the soft prompt distribution generated by the prompter. In WSV, the input received by GPT-2 is no longer a conventional discrete text prefix, but a sequence of visual prompt embeddings derived from polished video latents. Therefore, updating GPT-2 helps the decoder better interpret these prompts and convert them into more accurate captions.

The improvement on CIDEr further suggests that fine-tuning strengthens the generation of content words that are more consistent with human annotations, while the gains on B@4 and ROUGE-L show better n-gram precision and sentence-level overlap. These results demonstrate that the language model is not merely a passive text generator in WSV. Instead, it is an important part of the visual-to-language alignment process. Therefore, we fine-tune GPT-2 together with the prompter in the final configuration.

\begin{table}[t]
    \centering
    \caption{Ablation Study of Polisher Structure on MSR-VTT Dataset}
    
    \begin{tabular}{ccccc}
    \toprule
        \multirow{2}{*}{Method} & \multicolumn{4}{c}{MSR-VTT} \\
        \cmidrule(lr){2-5}
        & B@4 & M & R & C\\
        \midrule
        Wan2.2-T2V & 33.6 & \textbf{27.9} & \textbf{56.0} & 44.8\\
        CogVideoX & \textbf{33.9} & 27.7 & 55.7 & \textbf{45.5}\\
    \bottomrule
    \end{tabular}
    
    \label{tab:t2v}
\end{table}

\textbf{The Impact of Different T2V Model.} Table~\ref{tab:t2v} compares the effect of using different text-to-video models for visual latent synthesis. Since the T2V model provides the synthetic visual distribution used in the first training stage, its generated latent quality directly affects the subsequent polisher, prompter, and language model. The results show that both Wan2.2-T2V~\cite{wan2025wan} and CogVideoX can provide effective synthetic visual representations for WSV. Wan2.2-T2V achieves 33.6 on B@4, 27.9 on METEOR, 56.0 on ROUGE-L, and 44.8 on CIDEr, while CogVideoX achieves 33.9, 27.7, 55.7, and 45.5 on the same metrics.

The two T2V models show very close overall performance, but their advantages appear on different metrics. Wan2.2-T2V obtains slightly higher METEOR and ROUGE-L scores, exceeding CogVideoX by 0.2 and 0.3, respectively. This suggests that the captions generated from Wan2.2-T2V latents have competitive word-level semantic matching and sentence-level overlap. In contrast, CogVideoX achieves the best B@4 and CIDEr scores, with gains of 0.3 and 0.7 over Wan2.2-T2V. Since CIDEr is highly related to the consistency between generated captions and human reference descriptions, the higher CIDEr score indicates that CogVideoX provides visual latents that are slightly more favorable for generating discriminative and reference-consistent descriptions.

These results demonstrate that WSV is not tied to a single T2V backbone. As long as the T2V model can synthesize semantically meaningful and temporally coherent latent representations, the proposed framework can use them as visual surrogates for text-only training. At the same time, the small but consistent advantage of CogVideoX on B@4 and CIDEr shows that the compatibility between the generated latent space, the VAE encoder used at inference, and the polisher remains important. Therefore, we use CogVideoX as the default T2V model in WSV, while the comparison with Wan2.2-T2V further verifies the generality of the proposed visual synthesis strategy.

\begin{table}[t]
    \centering
    \caption{Ablation Study of Polisher Structure on MSR-VTT Dataset}
    
    \begin{tabular}{ccccc}
    \toprule
        \multirow{2}{*}{Method} & \multicolumn{4}{c}{MSR-VTT} \\
        \cmidrule(lr){2-5}
        & B@4 & M & R & C\\
        \midrule
        Transformer & 32.2 & 27.0 & 54.7& 42.9\\
        3D CNN & \textbf{33.9} & \textbf{27.7} & \textbf{55.7} & \textbf{45.5}\\
    \bottomrule
    \end{tabular}
    
    \label{tab:ps}
\end{table}

\textbf{The Impact of Polisher Structure.} Table~\ref{tab:ps} studies the influence of different polisher architectures. The polisher is responsible for refining synthetic video latents and reducing the distribution mismatch between generated videos and real videos. Therefore, its structure should preserve spatio-temporal information while correcting local artifacts in the synthesized latent representation. We compare a Transformer-based polisher with the 3D CNN structure used in WSV.

The 3D CNN polisher achieves better results on all metrics. Compared with the Transformer structure, it improves B@4 from 32.2 to 33.9, METEOR from 27.0 to 27.7, ROUGE-L from 54.7 to 55.7, and CIDEr from 42.9 to 45.5. The improvement of 2.6 on CIDEr is the most significant, indicating that the 3D CNN structure produces refined latents that help the captioner generate more informative and reference-consistent descriptions. The gains on B@4, METEOR, and ROUGE-L further show that the improvement is not limited to a single metric, but appears across lexical precision, semantic matching, and sentence-level overlap.

The superiority of the 3D CNN structure is consistent with the role of the polisher in WSV. The synthetic latent representation has a natural spatio-temporal organization, and 3D convolution can directly model local correlations across channels, frames, height, and width. This inductive bias is well suited for removing local noise and enhancing motion-related visual patterns while preserving the original latent layout. By contrast, a Transformer-based structure usually needs to flatten or tokenize the latent representation, which may weaken local continuity and make the refinement process less stable for dense video latents. Therefore, the 3D CNN polisher provides a more appropriate structure for latent-level video refinement, and we adopt it as the final polisher architecture.

\section{Conclusion}
\label{sec:Conclusions}

In this paper, we propose WSV to address the modality gap between training (text-only) and inference (video) in zero-shot video captioning. Previous methods align modalities using simple linear transformations in the latent space, resulting in insufficient alignment and a persistent, non-negligible gap between modalities. In contrast, we propose using a state-of-the-art text-to-video generation model and a polisher during training to conform text representations to a fundamental visual distribution directly, making the video distribution visible during training, thus fundamentally addressing the modality gap between vision and language. Furthermore, we propose a prompter to generate prompts for GPT-2 based on the visual latent representation, and finally utilize GPT-2 to generate the caption. Extensive experiments on MSVD, MSR-VTT and VATEX demonstrate that our method achieves excellent zero-shot video captioning performance. Future work will explore more powerful generation models and lightweight versions to enhance efficiency.


\bibliographystyle{IEEEtran}
\bibliography{references}

\end{document}